\documentclass[letterpaper, 10 pt, conference]{ieeeconf}  

\IEEEoverridecommandlockouts                              

\usepackage{times}
\usepackage{multicol}
\usepackage[bookmarks=true]{hyperref}
\usepackage{amsfonts}
\usepackage{amssymb}
\usepackage{amsmath}
\usepackage{graphicx}
\usepackage{color,soul}
\usepackage[ruled,vlined]{algorithm2e}
\usepackage{xcolor}
\usepackage{subcaption}
\usepackage{paralist}
\usepackage{siunitx}
\usepackage{mathtools}
\newtheorem{theorem}{Theorem}
\newtheorem{lemma}{Lemma}

\newtheorem{definition}{Definition}
\newtheorem{remark}{Remark}

\title{\LARGE \bf
	Multi-robot energy autonomy with wind and constrained resources
}

\author{Hassan Fouad and Giovanni Beltrame
\thanks{*This work was supported by the NSERC Discovery Grant 2019-05165.}
\thanks{Hassan Fouad and Giovanni Beltrame are with Computer and Software Engineering department, Ecole polytechnique de Montreal {\tt\small hassan.fouad@polymtl.ca}}%
\thanks{**The appendices of this paper can be found at \url{https://mistlab.ca/papers/energy2020}}
}

\begin{document}

\maketitle
\thispagestyle{empty}
\pagestyle{empty}

\begin{abstract}
  One aspect of the ever-growing need for long term autonomy of
  multi-robot systems, is ensuring energy sufficiency. In particular,
  in scenarios where charging facilities are limited, battery-powered
  robots need to coordinate to share access.  In this work we extend
  previous results by considering robots that carry out a generic
  mission while sharing a single charging station, while being
  affected by air drag and wind fields. Our mission-agnostic framework
  based on control barrier functions (CBFs) ensures energy sufficiency
  (i.e., maintaining all robots above a certain voltage threshold) and
  proper coordination (i.e., ensuring mutually exclusive use of the
  available charging station). Moreover, we investigate the
  feasibility requirements of the system in relation to individual
  robots' properties, as well as air drag and wind effects. We show
  simulation results that demonstrate the effectiveness of the
  proposed framework.
\end{abstract}

\section{Introduction}
The continuous advances in multi-robot systems gave rise to many new
applications like patrolling~\cite{portugal2011survey},
coverage~\cite{cortes2004coverage},
exploration~\cite{burgard2005coordinated} and
construction~\cite{petersen2011termes} to give a few examples. This
has drawn many researchers' attention in recent years to long term
autonomy and resilience of multi-robot systems, with the aim of
providing more practical and robust systems.

Energy autonomy, the ability of the robots in a multi-robot system to
replenish their energy reserves, is particularly important to extend
mission duration and general survivability.

Earlier interest in optimizing energy consumption in a multi-robot
system can be traced back to energy aware path
planning~\cite{sun2003energy} and node scheduling in wireless sensor
networks~\cite{slijepcevic2001power}. These ideas have been applied to
multi-robot systems as in~\cite{sonia}, where the mission tasks are
divided among robots according to their energy content.

One option for tackling the issue of limited energy is through the
introduction of stationary or mobile charging stations. Ding et
al.~\cite{ding2019decentralized} propose a method for planning routes
of charging robots that deposit batteries along the trajectories of
other robots carrying out a surveillance mission. Notomista et
al.~\cite{notomista2018persistification} use a control barrier
function framework that allows each robot in a multi-robot system to
recharge from a dedicated static charging station in a mission
agnostic and minimally invasive manner.

%
%
%
%
%

In this work we extend~\cite{fouadEnergy2020}, which is in turn
inspired from~\cite{notomista2018persistification}, by considering a
group of robots affected by air drag that perform a generic mission
(e.g. coverage or patrolling) in a known wind field. These robots need
to share a single charging station.

The contributions of this paper are: 1) We extend the results
in~\cite{fouadEnergy2020} so that the CBF-based coordination framework
proposed can account for the effect of air drag and winds, while
ensuring mutually exclusive use of the charging station, and 2) we
extend the sufficient feasibility conditions proposed
in~\cite{fouadEnergy2020} to express the system's capacity in case of
wind and air drag effects and ensure the feasibility of coordination.

\section{Preliminaries \label{sec:prelim}}
\subsection{Control Barrier Functions (CBF)}
A control barrier function (CBF) ~\cite{ames2019control} is a tool
that is mainly used to ensure set invariance of control affine
systems, having the form
\begin{equation}
\dot{x} = f(x)+g(x)u.
\label{eqn:control_affine}
\end{equation} 
This is often used for ensuring system's safety by enforcing forward
invariance of a desired safe set.

The safe set is defined to be the superlevel set of a continuously
differentiable function $h(x)$ such that~\cite{ames2019control}:
\begin{equation}
\begin{split}
\mathcal{C} &= \{x\in\mathbb{R}^n:h(x)\geq 0\}\\
\partial\mathcal{C} &= \{x\in\mathbb{R}^n:h(x)=0\}\\
Int(\mathcal{C}) &= \{x\in\mathbb{R}^n:h(x)>0\}\\
\end{split}
\end{equation}
where ensuring $h(x)>0,\forall t\geq 0$ implies the safe set $\mathcal{C}$ is
positively invariant. For a control affine system, having a control action $u$ that achieves 
\begin{equation}
\underbrace{L_fh(x)+L_gh(x)u}_{\dot{h}(x)} \geq -\alpha(h(x)) 
\label{eqn:basic_constraint}
\end{equation}
where $\alpha(h(x))$ is an extended type $\mathcal{K}$ function,
ensures positive invariance of $\mathcal{C}$.

One popular type of CBFs that we use in this paper is the zeroing
control barrier function (ZCBF)~\cite{ames2016control}, as they have
favourable robustness and asymptotic stability
properties~\cite{xu2015robustness}.
\begin{definition}{~\cite{ames2016control}}
  For a region $\mathcal{D}\in\mathcal{C}$ a continuously
  differentiable function $h(x)$ is called a ZCBF if there exists an
  extended class $\mathcal{K}$ function $\alpha(h(x))$ such that
  \begin{equation}
    \sup_{u\in U}\left(L_fh(x)+L_ghu+\alpha(h(x))\geq 0\right)
  \end{equation}
\end{definition}
The set $K_{zcbf}$\cite{ames2016control} is defined as
\begin{equation*}
  K_{zcbf}=\{u\in U:L_fh(x)+L_ghu+\alpha(h(x))\geq 0\}
\end{equation*}
and it is the set that contains all the safe control inputs, thus
choosing a Lipschitz continuous controller $u\in K_{zcbf}$ ensures
forward invariance of $\mathcal{C}$ and system's safety.

To mix the safety control input with an arbitrary mission's control
input $u_{nom}$, we use a quadratic program:
~\cite{notomista2018persistification}
\begin{equation}
\begin{aligned}
& u^*=\underset{u}{\text{min}}
& & ||u-u_{nom}||^2 \\
& \quad\quad\quad \text{s.t.}
& & L_fh(x)+L_gh(x)u \geq -\alpha(h(x)).
\end{aligned}
\label{eqn:QP}
\end{equation}

\subsection{Higher order control barrier functions (HOCBF)}
If $h(x)$ is of a higher relative degree (the control action $u$
doesn't appear after differentiating once, i.e. $L_gh(x)=0$),
using~\eqref{eqn:basic_constraint} to find an appropriate control
action becomes invalid. HOCBFs~\cite{xiao2019control} are an effective
solution of this problem. To define a HOCBF, we first need to define
the following set of functions for an $m^{th}$ order differentiable
function $h(x)$
\begin{equation}
\begin{split}
\psi_0(x) &= h(x)\\
\psi_1(x) &= \dot{\psi}_0(x) + \alpha_1(\psi_0(x))\\
\vdots\\
\psi_m(x) &= \dot{\psi}_{m-1}(x) + \alpha_{m}(\psi_{m-1}(x))
\end{split}
\label{eqn:psi}
\end{equation}
where $\alpha_1,\dots,\alpha_{m-1}$ are class $\mathcal{K}$ functions. Also we define the following series of sets 
\begin{equation}
\begin{split}
\mathcal{C}_1 \vcentcolon&=\{x\in\mathbb{R}^n: \psi_0(x) \geq 0\}\\
\vdots\\
\mathcal{C}_m \vcentcolon&=\{x\in\mathbb{R}^n: \psi_m(x) \geq 0\}
\end{split}
\label{eqn:c}
\end{equation}
\begin{definition}{\cite{xiao2019control}}
  Let $\mathcal{C}_1,\mathcal{C}_2,\dots,\mathcal{C}_m$ be defined by
  \eqref{eqn:c} and $\psi_0(x),\psi_1(x),\dots,\psi_m(x)$ be defined
  by \eqref{eqn:psi}. A function $h(x)$ is a HOCBF of relative degree
  $m$ for system~\eqref{eqn:control_affine} if there exists
  differentiable class $\mathcal{K}$ functions
  $\alpha_1,\alpha_2,\dots,\alpha_m$ such that
	\begin{equation}
	L_{f}^{m}h(x)+L_gL_{f}^{m-1}h(x)+O(h(x))+\alpha_{m}(\psi_{m-1}(x))\geq 0
	\label{eqn:hocbf}
	\end{equation}
	for all
        $x\in \mathcal{C}_1\cap
        \mathcal{C}_2\dots\cap\mathcal{C}_m$. Here $O(h(x))$ denotes
        the remaining Lie derivatives along $f$ with degrees less than
        or equal $m-1$.
\end{definition}
Xiao et al. show in~\cite[Theorem 5]{xiao2019control} that choosing a
control action that satisfies~\eqref{eqn:hocbf} renders the set
$\mathcal{C}_1\cap \mathcal{C}_2\dots\cap\mathcal{C}_m$ forward
invariant for system~\eqref{eqn:control_affine}.

\section{Problem formulation}
We assume $n$ robots moving in a given wind field with the following
dynamics:
\begin{equation}
\begin{split}
\dot{\mathbf{x}} &= \mathbf{v}\\
\dot{\mathbf{v}} &= u - C_d(\mathbf{v} - \mathbf{v}_w)\\
\dot{E} &= \begin{cases}
-k_e - k_v||\mathbf{v} - \mathbf{v}_w|| \quad, \text{if} ||\mathbf{x}-\mathbf{x}_c|| > \delta\\
k_{ch}\quad ,\text{Otherwise}
\end{cases}
\end{split}
\label{eqn: system model}
\end{equation}
where $\mathbf{x} \in\mathbb{R}^2$ is the robot's position,
$\mathbf{v} \in\mathbb{R}^2$ is its velocity,
$C_d>0,k_{ch}>0,k_e>0,k_v>0$ are coefficients of linear drag,
recharge, static and dynamic discharge respectively. Also $E > 0$ is
the robot's voltage, $u \in \mathbb{R}^2$ is the control input (no
constraints on the control input), and $\mathbf{v}_w$ is a known wind
vector. Moreover we suppose that all the robots operate in a certain
known operational range $\mathbf{x}\in\mathcal{R}\subset\mathbb{R}^2$,
where $\mathcal{R}$ is a closed set, and the size of this operational
range is described by the operational radius $R_0$. The robots are
carrying out a mission specified by $u_{nom}$ and they have one
charging station at a known location $x_c$ (in the origin without
loss of generality), and this station can only serve one robot at a
time, and has an effective charging range of $\delta > 0$.

We point out that in our model we use a linear drag term to account
for the air drag effect, which is a reasonable approximation for
bodies moving at low speeds.  The main assumptions we are adopting in
this work are:
\begin{inparaenum}
\item all robots have the same properties
\item robots have a complete communication graph
\item robots start discharging from the maximum voltage
\item the charging rate is faster than the discharge rate\footnote{E.g. battery swapping or high power wireless charging.}
\item An upper bound of the average relative velocity w.r.t. wind velocity(we call it $\tilde{V}$) of all robots is known at the beginning of the mission.
\end{inparaenum}
We propose a CBF framework that:
\begin{itemize}
\item Ensures no robot runs out of energy during the mission
\item Coordinates the times of arrival to the charging station so they are mutually exclusive.
\end{itemize}

Additionally, we describe the system's capacity as the relationship
between number of robots and robot properties with feasible
separation in arrival times at the charging station.

\emph{It is worth mentioning that for the rest of the paper we are omitting the proofs due to space constraints, and putting them all in the appendix.}
\section{Energy sufficiency  \label{sec:energy_suff}}
We provide a CBF that ensures that the voltage of all robots does not
go below a certain desired minimum voltage $E_{min}$. We take
inspiration from~\cite{notomista2018persistification}, but we extend
it to accommodate the system dynamics in~\eqref{eqn: system
  model}. The candidate CBF is
\begin{equation}
h_e = E - E_{min}-k_c\log\frac{D}{\delta}
\label{eqn:hs}
\end{equation}
where $D = ||x-x_c||$\footnote{The choice of $k_c$ is explained in the
  appendix}. The first derivative of this function is
\begin{equation}
\dot{h}_e = -k_e-k_v||\mathbf{v} - \mathbf{v}_w||-\frac{k_v}{D^2}(\mathbf{x}-\mathbf{x}_c)^T\mathbf{v}
\end{equation}
so we need to differentiate twice for the control input $u$ to appear
\small
\begin{equation}
\begin{split}
  &\ddot{h}_e = \overbrace{-\left[kv\frac{(\mathbf{v} - \mathbf{v}_w)^T}{||\mathbf{v} - \mathbf{v}_w||}+k_c\frac{(\mathbf{x} - \mathbf{x}_c)^T}{D^2}\right]}^{L_gL_fh_e(x)}u\\
  &+k_vC_d||\mathbf{v} -
  \mathbf{v}_w||\\&+\frac{k_c}{D^2}\left[\frac{2((\mathbf{x}-\mathbf{x}_c)^T\mathbf{v})^2}{D^2}-
    \mathbf{v}^T\mathbf{v}
    +C_d\mathbf{x}-\mathbf{x}_c)^T(\mathbf{v}-\mathbf{v}_w) \right]
\end{split}
\end{equation}
\normalsize with the second and third expressions being
$L_{f}^{2}h_e(x)$. We can then create an inequality similar
to~\eqref{eqn:hocbf} using $\alpha_1(h) = p_1h$ and
$\alpha_2(h)=p_2h$
\begin{equation}
L_{f}^{2}h_e(x) + L_gL_fh_e(x)u +(p_1+p_2)\dot{h}_e+p_1p_2h_e\geq 0
\label{eqn:he_ineq}
\end{equation}
and $p_1>$ and $p_2>$ are chosen in such a way that lends the
characteristic equation of the left side of~\eqref{eqn:he_ineq} with
distinct real roots.
\begin{theorem}
  \label{thm:he_hocbf}
  For a robot described by dynamics in \eqref{eqn: system model}, and
  provided that the robot is out of the charging region, and that
  $k_c > k_vR_0$, then $h_e$ is a HOCBF.
\end{theorem}

\begin{lemma}
  \label{lemma:emin_arrival}
  For a robot with dynamics described by \eqref{eqn: system model}
  applying a QP as in \eqref{eqn:QP} with \eqref{eqn:he_ineq} as being
  the constraint, then the quantity $E-E_{min}$ at the time of arrival
  to the charging station is upper bounded with a quantity
  exponentially decaying with a rate of
  $\frac{1}{2}\left(-(p_1+p_2)+|p_1-p_2|\right)$ and lower bounded by zero.
\end{lemma}

\section{Coordination}
The second component in our framework ensures that the difference in
arrival times of any two robots to the charging station is above a
desired limit. The main idea is that if two robots have different
values of $E_{min}$, they arrive to the charging station at different
times. We propose a method for changing the values of $E_{min}$ to
achieve the aforementioned coordination.

To get this expression, we integrate the voltage relation
in~\eqref{eqn: system model} to get
\begin{equation}
\begin{split}
\int_{E_{max}}^{E_{min}}\dot{E} dt&= - \int_{0}^{T_L} (k_e+k_v||\mathbf{v}-\mathbf{v}_w||)dt\\
E_{max}-E_{min}&=k_eT_L+k_v\int_{0}^{T_L} ||\mathbf{v}-\mathbf{v}_w||dt.
\end{split}
\end{equation}
Supposing we have the average relative speed
$\bar{z}=\frac{1}{T_L}\int_{0}^{T_L}||\mathbf{v}-\mathbf{v}_w||dt$,
the last integral can be replaced and the arrival time becomes
\begin{equation}
T_L=\frac{E_{max}-E_{min}}{k_e+k_v\bar{z}}.
\end{equation}
We then replace $E_{max}$ in the last expression by $E(t)$ to get an
expression for $T_L(t)$ that changes with time
\begin{equation}
T_L(t)=\frac{E(t)-E_{min}}{k_e+k_v\bar{z}}.
\label{eqn:arrival_time}
\end{equation}
In this work, we use a moving average $\bar{V}$ to estimate the
average velocity relative to wind defined as
\begin{equation}
\bar{V} = \frac{1}{w}\int_{0}^{w}||\mathbf{v}-\mathbf{v}_w||dt
\end{equation} 
where $w>0$ is the width of the integration window. The larger the
window, the closer the estimate is to the true average. The
approximate value of the arrival time is
\begin{equation}
T_L(t)\approx\frac{E(t)-E_{min}}{k_e+k_v\bar{V}}.
\label{eqn:arrival_time_approx}
\end{equation}
To be able to change $E_{min}$ to achieve coordination, we propose a
simple single integrator model for $E_{min}$ as follows
\begin{eqnarray}
\dot{E}_{min}=\eta
\end{eqnarray}
where $\eta\in\Theta\subset\mathbb{R}$ is a control input to
manipulate $E_{min}$,and $\Theta$ is being the set of all possible
values of $\eta$.  It is useful to point out that $\eta$ has a default
value of $\eta_{nom}=0$ unless modified by the proposed coordination
framework.

\subsection{Coordination CBF}
We propose a CBF approach to change the values of $E_{min}$ to ensure
mutually exclusive use of the charging station. We define a
coordination CBF $h_{c_{ij}}$ between robots $i$ and $j$, as well as
an associated pairwise safe set $\mathcal{C}_{ij}$
\begin{equation}
  \mathcal{C}_{ij} = \{(E_{min_i},E_{min_j})\in\mathbb{R}^2 : h_{c_{ij}} \geq 0\}.
\end{equation}
We use the same coordination CBF as in~\cite{fouadEnergy2020}
\begin{equation}
h_{c_{ij}} = \log\frac{|T_{L_i}-T_{L_j}|}{\delta_t}
\label{eqn:coord_CBF}
\end{equation}
and to get a constraint similar to \eqref{eqn:basic_constraint}
\begin{equation}
\frac{T_{L_i}-T_{L_j}}{|T_{L_i}-T_{L_j}|^2}(\theta_i\eta_i+\beta_i-\beta_j)\geq \alpha(h_{c_{ij}})
\label{eqn:coord_constraint}
\end{equation}
where 
\small
\begin{equation*}
\begin{split}
&\theta_i = -\frac{1}{k_e+k_v\bar{V}}\\
&\beta_i =\frac{-k_e-k_vV}{k_e+k_v\bar{V}}-\frac{k_v}{w}\frac{(E-E_{min})(V(t)-V(t-w))}{k_e+k_v\bar{V}}\\
&\dot{T}_{L_i} = \theta_i\eta_i+\beta_i
\end{split}
\end{equation*}
\normalsize where $V = ||\mathbf{v}-\mathbf{v}_w||$. For decentralized
implementation, we dropped out the term $\theta_j$ so the constraint
equation is independent of $\eta_j$, and provided both robots are
adopting the constraint \eqref{eqn:coord_constraint}, each will try to
stay in the safe set $\mathcal{C}_{ij}$. For the right hand side of
\eqref{eqn:coord_constraint} we use the following
\begin{equation}
\begin{split}
&\alpha(h_{c_{ij}})=\gamma_{ij}.\text{sign}(h_{c_{ij}}).|h_{c_{ij}}|^\rho\quad, \rho\in\left[0,1\right)\\
&\gamma_{ij}=\begin{cases}
\gamma_h\quad,\text{if}\hspace{2mm} D_i>\delta \text{ and }D_j>\delta\\
0\quad,\text{otherwise}
\end{cases}
\end{split}
\label{eqn:gamma_choice}
\end{equation}
which is inspired from~\cite{li2018formally} and leads to the
favourable quality of converging to the safe set in a finite time, in
case the initial condition is out of the safe set.

\begin{theorem}{\cite[Theorem 2]{fouadEnergy2020}}
	\label{thm:hc}
	For a pair of robots $(i,j)$ that belongs to a multi-robot
        system and satisfying $D_i>\delta$ and $D_j>\delta$, and
        provided that $\eta\in\Theta=\mathbb{R}$ then $h_{c_{ij}}$ is
        a ZCBF. Moreover, if
        $(E_{min_i}(t_0),E_{min_j}(t_0))\notin\mathcal{C}_{ij}$, then
        the constraint \eqref{eqn:coord_constraint} leads
        $(E_{min_i}(t),E_{min_j}(t))$ to converge to
        $\partial\mathcal{C}_{ij}$ in finite time.
\end{theorem}

\subsection{Lower bound on $E_{min}$} 
Since $E_{min}$ is supposed to be the voltage at which the robot
arrives to the charging station, then it is necessary to enforce a
lower bound on its value to avoid any potential damage to the
batteries or the loss of a robot with excessively low voltage.
For this reason, we propose another CBF:
\begin{equation}
h_L = k_s(E_{min}-E_{lb})
\label{eqn:lb_cbf}
\end{equation}
where $E_{lb}>0$ is the desired lower bound voltage and $k_s>0$ is a
scaling gain. Differentiating $h_L$ and obtaining the QP constraint
gives
\begin{equation}
k_s\eta \geq -\alpha(h_L)
\label{eqn:hL_const}
\end{equation}
where $\alpha(h_L) = p_Lh_L$ for $p_L>0$. It can be easily shown that
$h_L$ is a ZCBF, since $\eta\in\Theta=\mathbb{R}$ (no constraint on
$\eta$) then there exists a control input $\eta$ that satisfies
\eqref{eqn:hL_const}.

\subsection{System capacity description}
To successfully apply the coordination CBF in a pairwise manner, the
value of the desired $\delta_t$ should be reasonable with respect to
individual robot's properties and the number of robots in the system
(e.g. we can't ask for $\delta_t$ that is longer than the total
discharge time of a battery). We consider the relation between the
robots' parameters, their number and the feasible limits on $\delta_t$
as being an expression of the system's capacity.

We propose a sufficient condition on the upper and
lower limits of $\delta_t$, in relation to properties like maximum and
minimum battery voltages, discharge and recharge rates, and the number
of robots in the system.
\begin{figure}[!htb]
	\centering
	\scalebox{0.65}{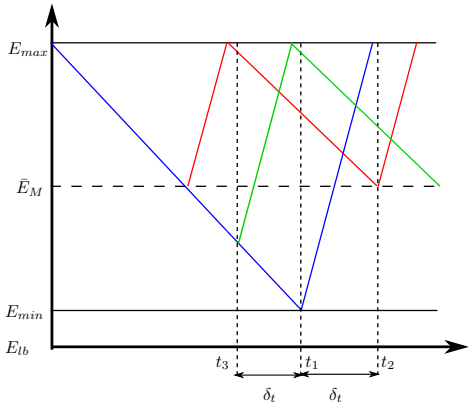}
	\caption{Schematic of a charging cycle with three robots. The red line represents the more needy robot, while the blue one is the least needy one, and it defines one recharging cycle.}
	\label{fig:capacity}
\end{figure}
For the sake of being conservative, we derive this capacity relation
assuming that the system is pushed to its limits, meaning that all
robots operate with the maximum average relative velocity w.r.t. wind
$\tilde{V}$. Suppose we have a group of $n$ robots, each has its own
$E_{min}$ value, and one of them is the ``neediest'' robot that
recharges first and most often, while another is the least needy one
(represented in Figure~\ref{fig:capacity} as the red and blue lines
respectively). We want $t_2-t_1\geq\delta_t$, which
means
{\small
\begin{equation}
\begin{split}
  &t_1 = \frac{E_{max}-E_{lb}}{k_e+k_v\tilde{V}}\\
  &t_2=\frac{E_{max}-\bar{E}_{M}}{k_e+k_v\tilde{V}}+\frac{E_{max}-\bar{E}_{M}}{k_{ch}}+\frac{E_{max}-\bar{E}_{M}}{k_e+k_v\tilde{V}}.
\end{split}
\end{equation}}

Calculating $t_2-t_1$ and considering that
$\bar{E}_M = E_M +\varepsilon$, where $E_M$ being \emph{the actual
  value of $E_{min}$ of the neediest robot during the coordination},
and $\varepsilon\geq 0$ being an additional increment of voltage to
$E_M$ that is caused by the dependence of discharge rate on
robot's speed, we have:
\begin{equation}
E_{M} \leq\frac{(1+\tfrac{k_e+k_v\tilde{V}}{k_{ch}})E_{max}+E_{lb}-\delta_t(k_e+k_v\tilde{V})-\kappa\varepsilon}{\kappa}
\label{eqn:em_ub}
\end{equation}
where $\kappa = 2+\tfrac{k_e+k_v\tilde{V}}{k_{ch}}$. We can then
calculate $\Delta E_M = \tfrac{E_M-E_{lb}}{n-1}$ which is a uniform
increment of $E_{min}$ between $E_{lb}$ and $E_M$ to create the
desired separation of arrival times
{\small
\begin{equation}
\Delta E_{M} = \frac{(1+\tfrac{k_e+k_v\tilde{V}}{k_{ch}})(E_{max}-E_{lb})-\delta_t(k_e+k_v\tilde{V})-\kappa\varepsilon}{\kappa(n-1)}
\label{eqn:delta_Emin}
\end{equation}}
What we want then is to have $t_1-t_3\geq\delta_t$, meaning that the
arrival times of the last two robots (or any two consecutive robots)
to be at least $\delta_t$
\begin{equation}
  \frac{E_{max}-E_{lb}}{k_e+k_v\tilde{V}} - \frac{E_{max}-(E_{lb}+\Delta E_{M})}{k_e+k_v\tilde{V}}\geq \delta_t
\end{equation} 
and substituting \ref{eqn:delta_Emin} into the last equation we get
\small
\begin{equation}
  \begin{split}
  &\frac{(1+\tfrac{k_e+k_v\tilde{V}}{k_{ch}})(E_{max}-E_{lb})-\delta_t(k_e+k_v\tilde{V})-\kappa\varepsilon}{\kappa(n-1)} \\&- \delta_t(k_e+k_v\tilde{V}) \geq 0
  \end{split}
\label{eqn:sys_capacity}
\end{equation}
\normalsize then we obtain a critical value of $\delta_t$ at which the
inequality becomes an equality
\begin{equation}
\delta_{t_{cr}} = \frac{\left(1+\frac{k_e+k_v\tilde{V}}{k_{ch}}\right)(E_{max}-E_{lb})-\kappa\varepsilon}{(k_e+k_v\tilde{V})\left[1+\kappa(n-1)\right]}.
\label{eqn:critical_dt}
\end{equation}

One final requirement on $\delta_{t_{cr}}$ is to be greater than half the time taken to recharge a battery from $E_{lb}$ to $E_{max}$
\begin{equation}
\delta_{t_{cr}} \geq \frac{E_{max}-E_{lb}}{2k_{ch}}
\label{eqn:lower_bnd_dt}
\end{equation}
The value of $\delta_{t_{cr}}$ represents in this case an upper bound
on the feasible $\delta_t$ that can be achieved by the system. To
motivate the need for $\varepsilon$\footnote{\label{note} More details on its
  derivation can be found in the appendix}, we consider the critical
case when $(E_{min_i},E_{min_j})\notin\mathcal{C}_{ij}$, in which case
the QP produces $\eta_i$ that renders \eqref{eqn:coord_constraint} an
equality, thus $\dot{h}_{c_{ij}}=\alpha(h_{c_{ij}})$ that reaches
steady state in finite time (i.e. coordination achieved) at which
$\dot{h}_{c_{ij}}=0$. Considering the case where all robots have same
$\bar{V}=\tilde{V}$ (which we already supposed when deriving
$\delta_{t_{cr}}$) then from the LHS of \eqref{eqn:coord_constraint}
we have $\eta_i=k_v(V_j-V_i)$. So if $V_i$ decreases (robot $i$ going
to recharge for example), $\eta_i$ increases and so
$\varepsilon=\Delta E_{min_i}$ resulting from the increase in
$\eta$. The value of $\varepsilon$ can be estimated by approximating
the integration of $\eta_i$ over the time it takes the robot to go
back to the charging station. One approximation for $\varepsilon$ is
\makeatletter
\newcommand\footnoteref[1]{\protected@xdef\@thefnmark{\ref{#1}}\@footnotemark}
\makeatother
\begin{equation}
\varepsilon = \frac{k_vV_n\left(T_{end}-\tfrac{n(E_{max}-E_{lb})-k_c\log\tfrac{R_0}{\delta}(1+\kappa(n-1))}{(k_e+k_v\tilde{V})(1+\kappa(n-1))}\right)}{1+\tfrac{k_vV_n}{k_e+k_v\tilde{V}}\left(1-\tfrac{1}{1+\kappa(n-1)}\right)}
\label{eqn:f_parameter}
\end{equation}
where
$T_{end} =
\frac{E_{max}-\tfrac{E_{max}+E_{lb}}{2}}{k_e+k_v\tilde{V}}$, and $V_n$
is the magnitude of the mission's nominal velocity w.r.t. the wind
velocity vector\footnoteref{note}.


\begin{lemma}
	\label{lemma:upper_limit_recharge}
	For a group of $n$ robots that have distinct values of
        $E_{min}$ that satisfy (\ref{eqn:critical_dt}),
        (\ref{eqn:lower_bnd_dt}) and (\ref{eqn:delta_Emin}), and
        provided they all operate such that their average relative
        velocity (w.r.t. wind) is equal to its upper bound,
        i.e. $\bar{V}=\tilde{V}$, let $z_i$ be the number of recharges
        that one robot can have in one charging cycle, then the
        maximum number of recharges for any robot is $\bar{z}_i = 2$. Moreover, $\bar{E}_M \leq \tfrac{E_{max}+E_{lb}}{2}$.
	\label{lemma:upper_bound_rechargeing_times}
\end{lemma}

\begin{lemma}
  \label{lemma:feasibility}
  For a group of $n$ robots, if $\delta_t$ satisfies
  \begin{equation}
    \frac{E_{max}-E_{lb}}{2k_{ch}} \leq \delta_t \leq \delta_{t_{cr}}
    \label{eqn:capacity}
  \end{equation}
  as well as equation (\ref{eqn:delta_Emin}), then there exists
  $\mathbf{E}_m=\{E_{min_1},\dots, E_{min_n}\}$ such that the
  difference in arrival times between any two robots is at least
  $\delta_t$ (i.e. the scheduling problem is feasible).
\end{lemma}

\subsection{Feasibility of QP}
In the proposed coordination framework so far $E_{min}$, which is a
1-D value, is being manipulated to vary the arrival times of robots to
the charging station. However, this can potentially cause a QP
infeasibility problem. For example, a robot might need to use a
negative $\eta$ to evade a neighbour's arrival time, but at the same
time it may need $\eta$ to be positive so as not to go below
$E_{lb}$. Some methods have been proposed to deal with this issue as
in ~\cite{wang2016multi} and~\cite{egerstedt2018robot}, and we adapt
the core idea of the latter. To avoid the infeasibility problem, each agent carries out
coordination only with its neighbour with the closest arrival
time. Moreover, it gives higher priority for maintaining
$E_{min}\geq E_{lb}$ over coordination. This way $\eta$ has to change
to adapt one thing at a time and avoid potential infeasibility (see Algorithm~\ref{algo}).

\begin{algorithm}[!htb]
	\SetAlgoLined
	\KwIn{$T_{L_k}\quad ,\forall k\in\mathcal{N}_i$}
	\KwResult{$A_c$ and $B_c$ for QP constraints}
	$h_{c_{min}}=h_0$
	
	$h_{L_i}=E_{min_i}-E_{lb}$
	
	\While{j in $\mathcal{N}_i$}{
		$h_{c_{ij}} = \log\tfrac{T_{L_i}-T_{L_j}}{\delta}$
		
		\If{$h_{c_{ij}}<h_{c_{min}}$ and $D_j>\delta$}{
			$h_{c_{min}}=h_{c_{ij}}$
		}
	}
	
	\eIf{$h_{c_{min}}<h_{L_i}$}{
		$A_c = L_g h_{c_{min}}$
		
		$B_c = -L_f h_{c_{min}}-\alpha(h_{c_{min}})$ \dots(eqn. \ref{eqn:coord_constraint})
	}{
	$A_c = L_g h_{L_i}$
	
	$B_c = -L_f h_{L_i}-\alpha(h_{L_i})$ \dots(eqn. \ref{eqn:hL_const})
}

\caption{Coordination algorithm}
\label{algo}
\end{algorithm}

The final QP is
\begin{equation}
\begin{aligned}
& \mathbf{u}^*=\underset{\mathbf{u}\in\mathbb{R}^3}{\text{min}}
& & ||\mathbf{u}-\mathbf{u}_{nom}||^2 \\
& \quad \quad\quad\text{s.t.}
& & A\mathbf{u} \geq B \\
\end{aligned}
\label{eqn:QP2}
\end{equation}
where 
\begin{equation*}
\begin{split}
A &= \begin{bmatrix}
A_e^T\\A_c^T
\end{bmatrix}=\begin{bmatrix}
L_gL_fh_e&0\\
 \mathbf{0}_{1\times 2} &A_c^T
\end{bmatrix},\\ B &= \begin{bmatrix}
B_e\\B_c
\end{bmatrix}=\begin{bmatrix}
-L_f^2h_e-(p_1+p_2)\dot{h}_e-p_1p_2h_e\\B_c
\end{bmatrix}
\end{split}
\end{equation*}
while $A_c$ and $B_c$ are determined from Algorithm~(\ref{algo}). 

\begin{theorem}~\cite[Theorem 3]{fouadEnergy2020}
	\label{thm:main_result}
        For a multi robot system of $n$ robots, with dynamics defined
        in \eqref{eqn: system model}, and each robot applying energy
        sufficiency, coordination and lower bound constraints defined
        in \eqref{eqn:he_ineq}, \eqref{eqn:coord_constraint} and
        \eqref{eqn:hL_const}, and provided that the inequalities
        (\ref{eqn:em_ub}) and (\ref{eqn:capacity}) are satisfied, then
        Algorithm~(\ref{algo}) ensures mutual exclusive use of the
        charging station.
\end{theorem}

\begin{figure*}[!tbp]
	\centering
	\begin{subfigure}[b]{0.32\textwidth}
		\includegraphics[width=0.9\columnwidth]{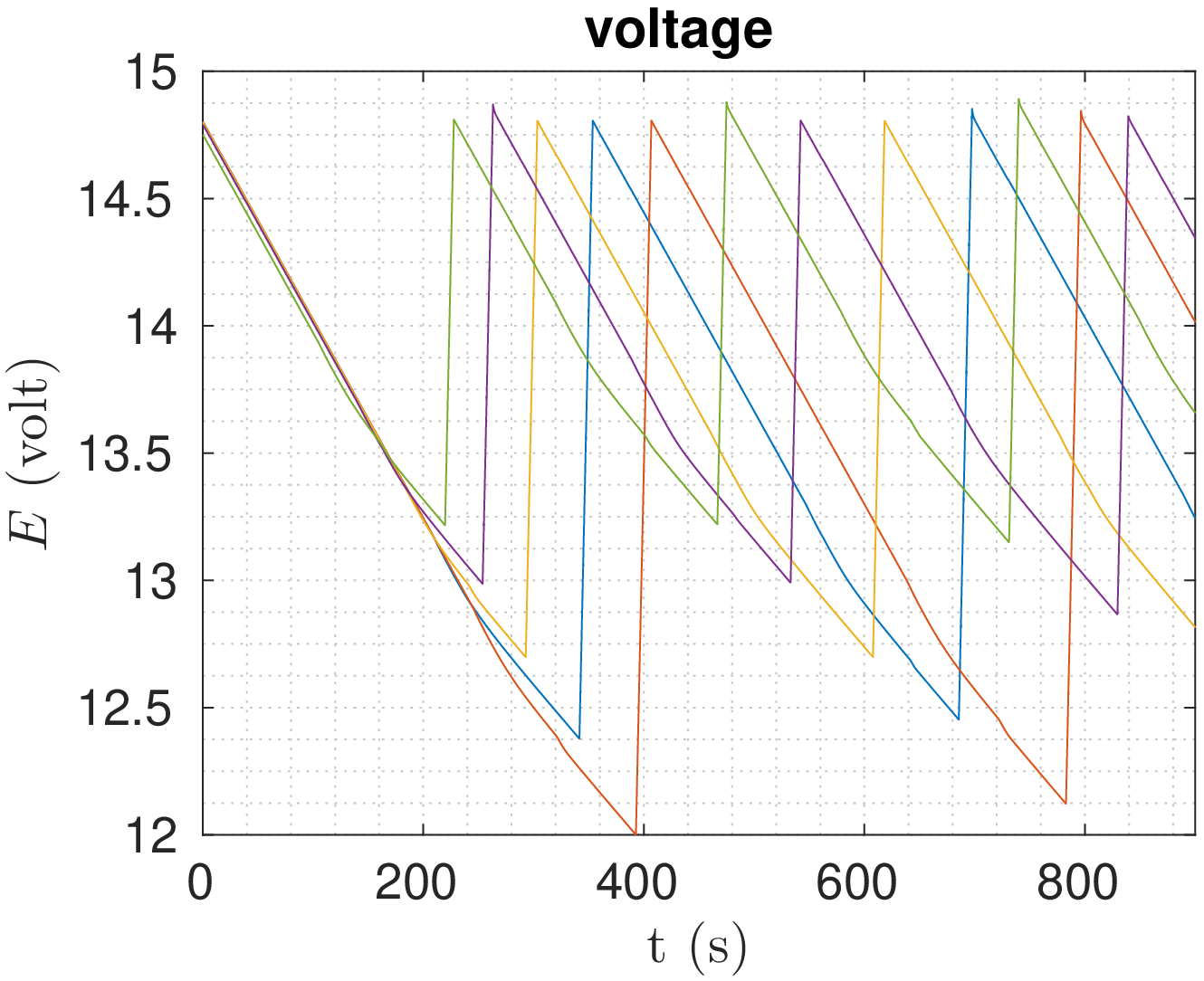}%
		\caption{}%
		\label{fig:volt}%
	\end{subfigure}
	\begin{subfigure}[b]{0.32\textwidth}
		\includegraphics[width=0.9\columnwidth]{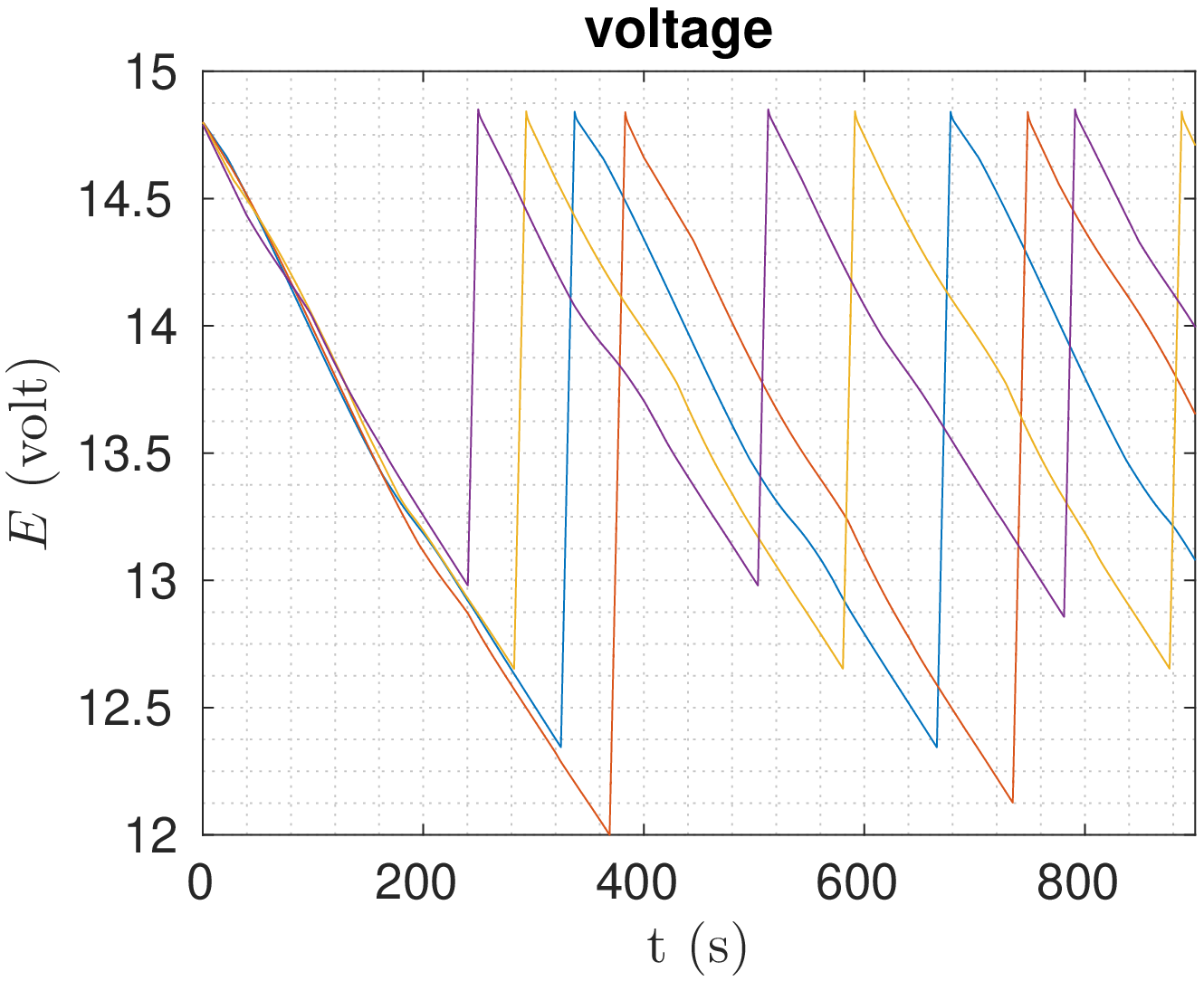}%
		\caption{}%
		\label{fig:volt2}%
	\end{subfigure}
	\begin{subfigure}[b]{0.32\textwidth}
		\includegraphics[width=0.9\columnwidth]{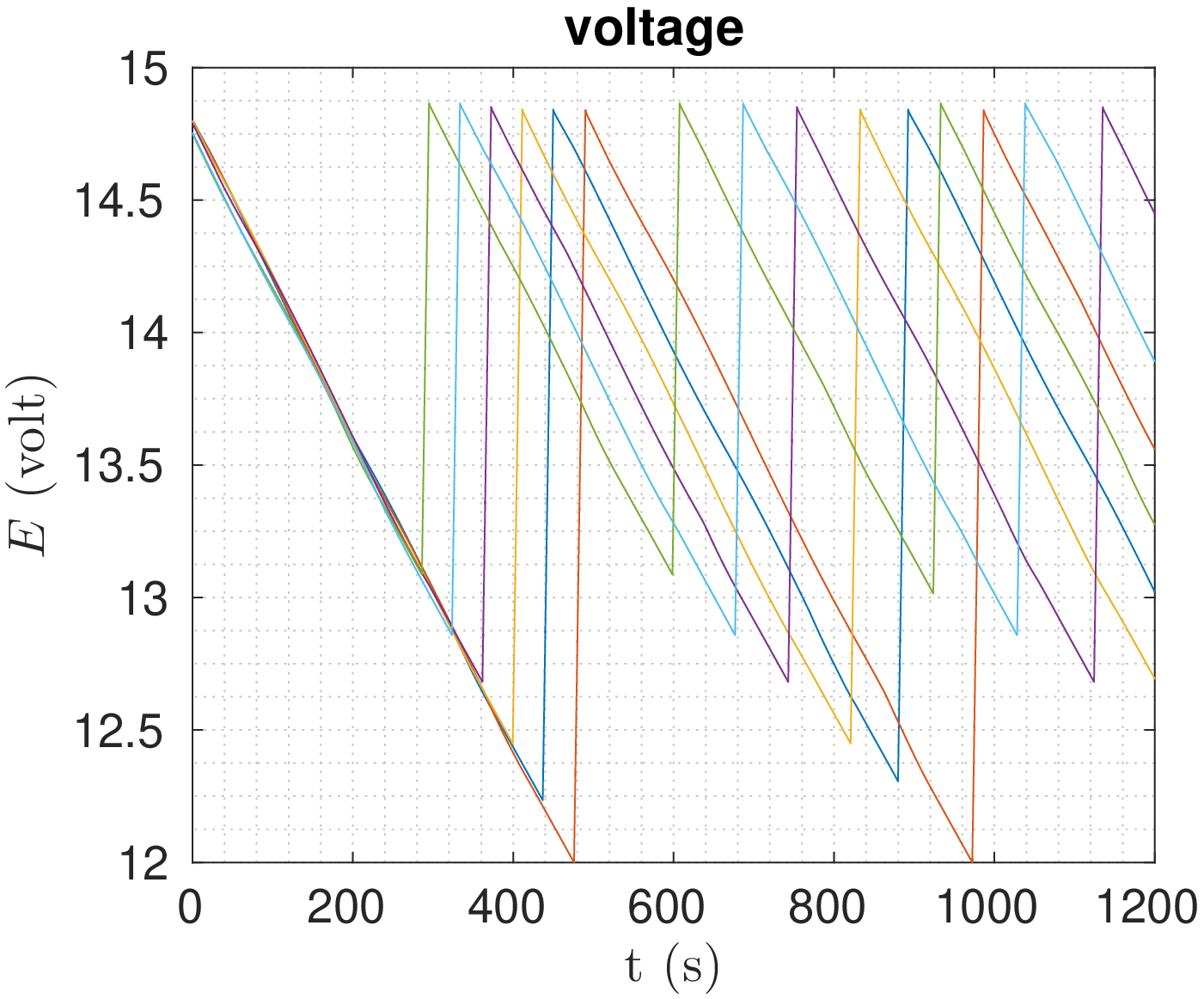}%
		\caption{}%
		\label{fig:volt3}%
	\end{subfigure}\hfill
	\begin{subfigure}[b]{0.32\textwidth}
		\includegraphics[width=0.9\columnwidth]{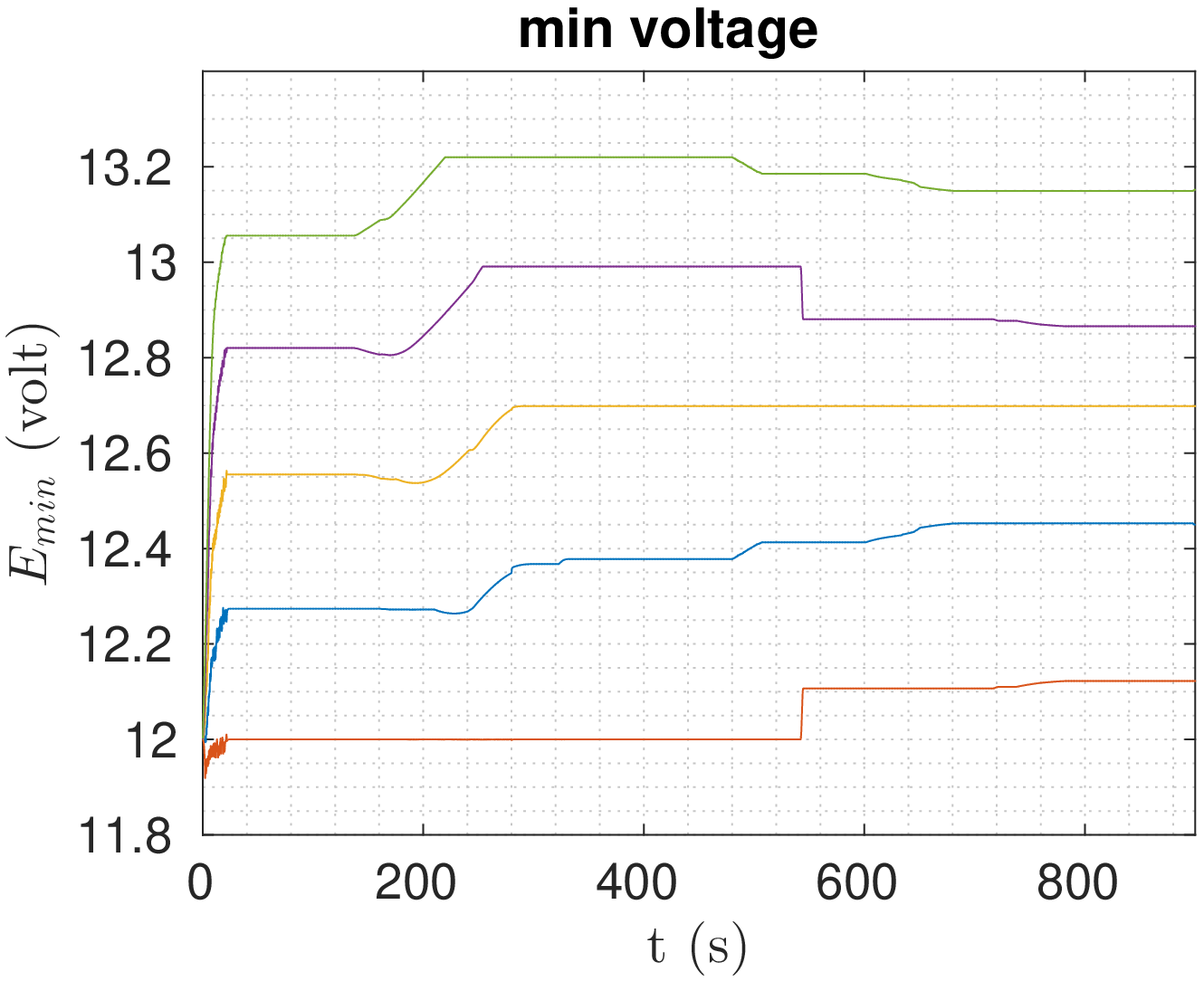}%
		\caption{}%
		\label{fig:emin1}%
	\end{subfigure}
	\begin{subfigure}[b]{0.32\textwidth}
		\includegraphics[width=0.9\columnwidth]{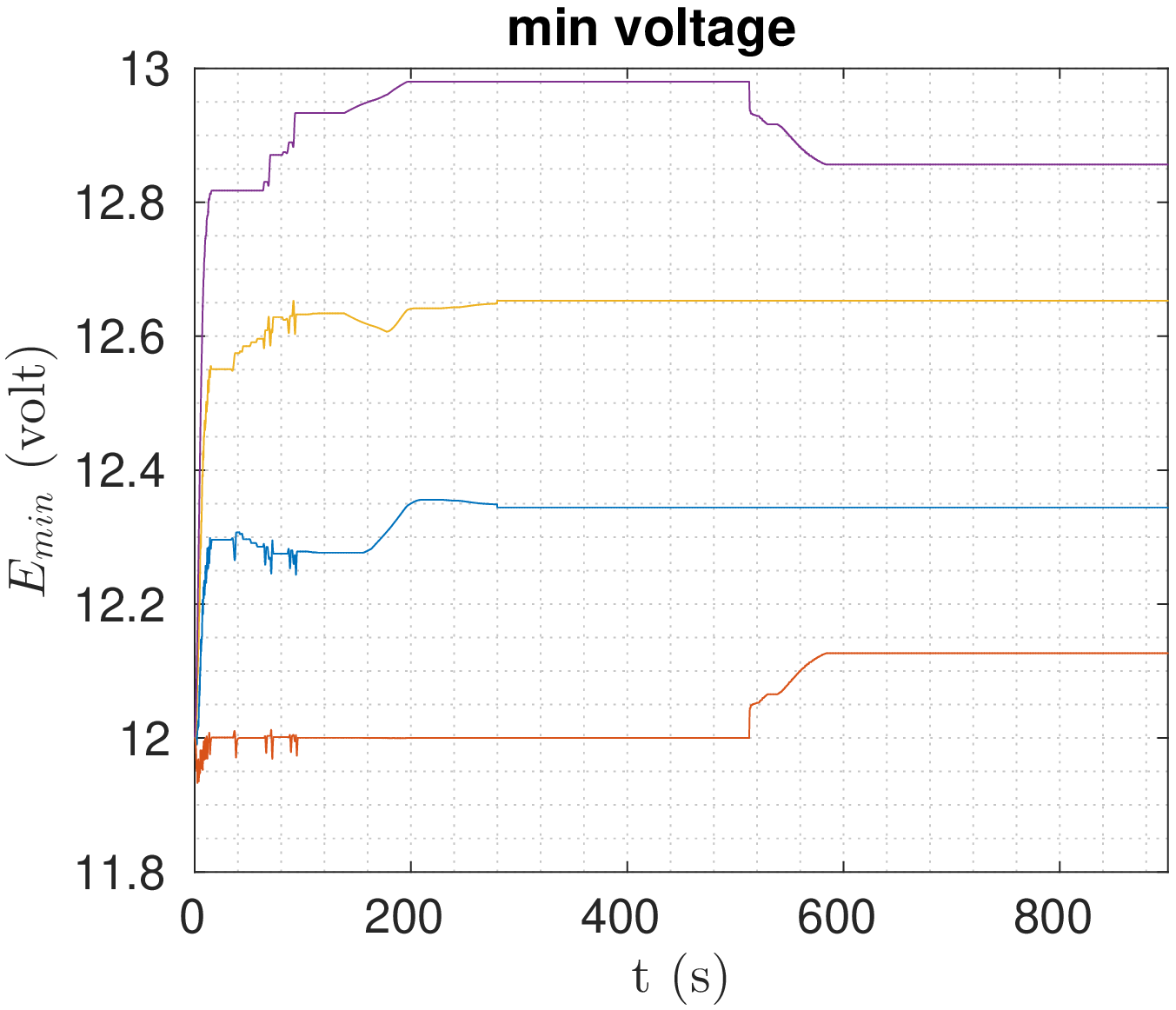}%
		\caption{}%
		\label{fig:emin2}%
	\end{subfigure}%
	\begin{subfigure}[b]{0.32\textwidth}
		\includegraphics[width=0.9\columnwidth]{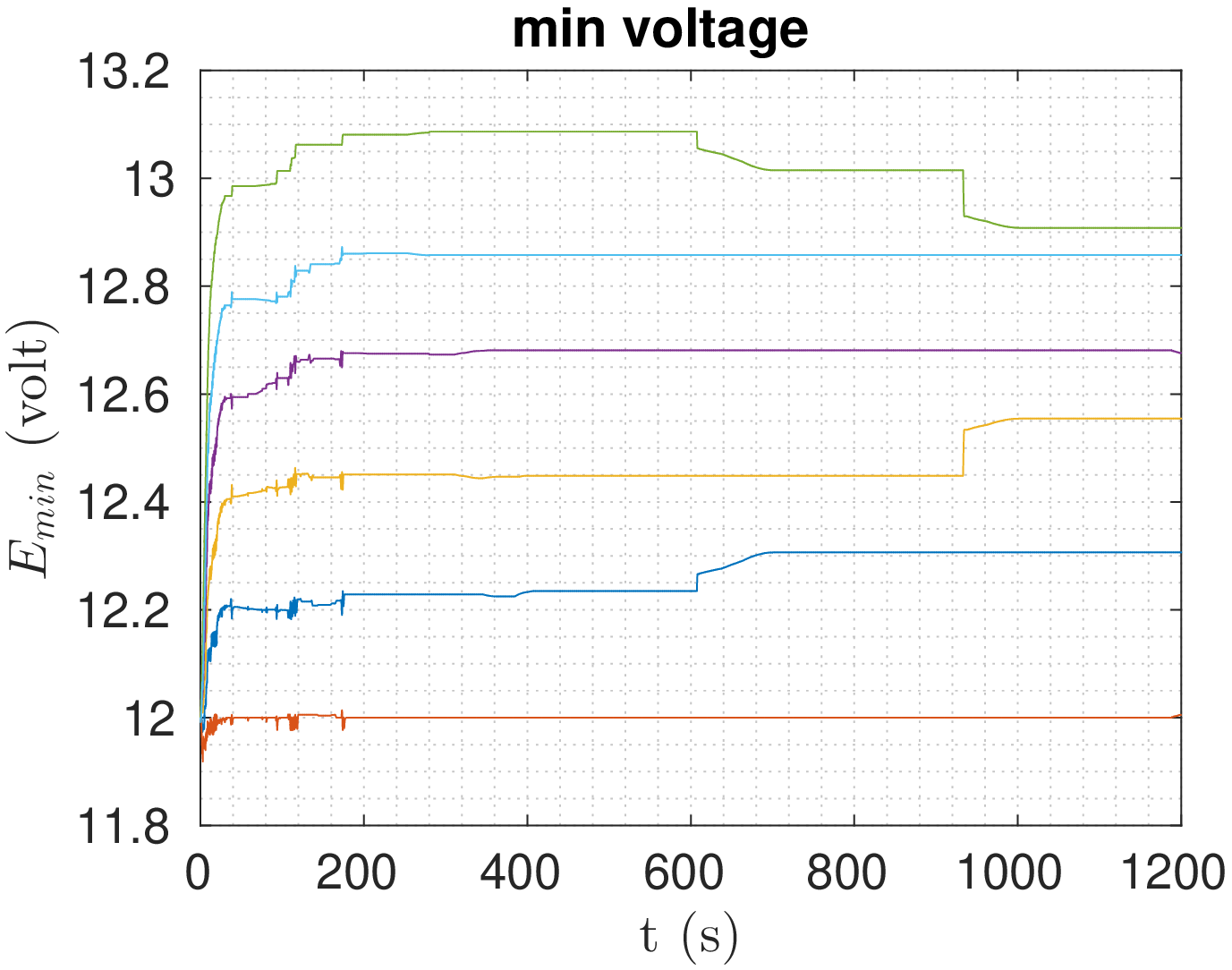}%
		\caption{}%
		\label{fig:emin3}%
	\end{subfigure}%
	
	\caption{Evolution of values of $E(t)$ and $E_{min}(t)$ for the three different scenarios under consideration}
	\label{fig:res1}
\end{figure*}
\section{Results \label{sec:results}}
In this section we present Matlab simulation results of the proposed
framework, aiming to highlight its effectiveness, as well as the
utility of the capacity estimation we propose. We tackled three
different scenarios of a simple patrolling mission for a group of robots
spinning around the charging station at a certain distance with a
desired nominal mission speed. In these scenarios we have
$\delta = 0.2\si{\meter}$, $R_0 = 9\si{\meter}$ and we want to achieve
$\delta_t = 35\si{\second}$. The results are depicted in
Figure~\ref{fig:res1}, and the main parameters used are presented in
Table~\ref{tab:params}. For all the cases discussed, the $\varepsilon$ is calculated using \eqref{eqn:f_parameter}.

\small
\begin{table}[!htb]
	\centering
	\caption{Values of parameters used in simulation}
	\label{tab:params}
	\begin{tabular}{c c c c c c}
		\hline 
		Parameter & $k_e$& $k_v$ & $k_{ch}$  & $E_{max}$ & $E_{lb}$ \\\hline
		Value & 0.005\si{\volt/\second}&0.015\si{\volt/\meter}&0.2\si{\volt/\second}&14.8\si{\volt}&12\si{\volt}\\\hline		
	\end{tabular}
\end{table}
\normalsize
\subsection{Base scenario}
In this scenario we have a group of five robots that revolve around
the charging station, with an upper bound of average relative velocity
$\tilde{V}=0.15\si{\meter/\second}$.

Each robot applies a proportional control on the speed to produce a
nominal control input $u_{nom} = -k_d(\mathbf{v}-\mathbf{v}_n)$, where
$\mathbf{v}_n$ is a nominal mission velocity and $k_d>0$ is a gain. The value of $u_{nom}$
is the one that goes into the QP \eqref{eqn:QP2}. To generate
$\mathbf{v}_n$ for patrolling, we specify a desired magnitude
$V_n=||\mathbf{v}_n||$ then we use potential flow theory to specify
the direction. We calculate a potential function $\phi$ of a source
near the charging station, and of a vortex near the boundary
\begin{equation}
\phi=\begin{cases}
m\log D\quad ,\text{if }\hspace{0.1cm}D<\delta+\Delta_{tol}\\
\frac{m}{2\pi}\theta_p,\text{if }\hspace{0.1cm}D>R_0+\Delta_{tol}
\end{cases}
\end{equation}
where $m>0$, $\Delta_{tol}>0$, $\theta_p=\angle(\mathbf{x}-\mathbf{x}_c)$ and $\mathbf{v}_n = V_n\tfrac{\nabla\phi}{||\nabla\phi||}$.

The requirement is to have a $\delta_t = 35\si{\second}$. The value of
$\delta_{t_{cr}}$ from \eqref{eqn:critical_dt} is
$\delta_{t_{cr}} = 36.39\si{\second}$ for $\varepsilon = 0.24$, and
$\tfrac{E_{max}-E_{lb}}{2k_{ch}}=7\si{\volt}$, thus
\eqref{eqn:capacity} is satisfied. The evolution of voltages and
$E_{min}$ values is depicted in figures~\ref{fig:volt} and~\ref{fig:emin1}.

\subsection{Base scenario with wind}
\begin{figure}
	\centering
	\includegraphics[scale=0.5]{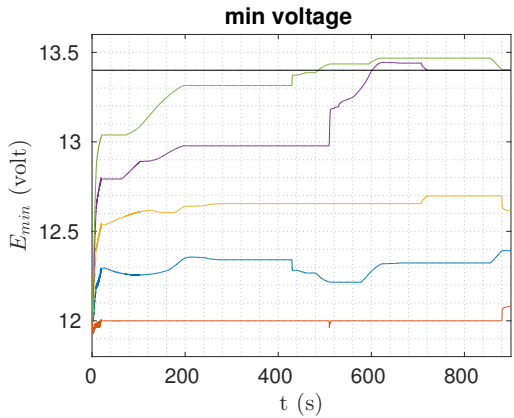}
	\caption{Plot of $E_{min}$ when \eqref{eqn:capacity} is not satisfied, as described in the second scenario.}
	\label{fig:bad_emin}
\end{figure}

Here we add a constant wind field of
$\mathbf{v}_w=(0.08,0.08)\si{\meter/\second}$ and we have an upper
bound $\tilde{V} = 0.2\si{\meter/\second}$. In this case
$\delta_{t_{cr}} = 31.9\si{\second}<\delta_t$ for
$\varepsilon = 0.28$. Choosing to use 5 robots causes $E_{min}$ for
some robots to go over $\tfrac{E_{max}+E_{lb}}{2}$ as shown in
Figure~\ref{fig:bad_emin} (when it should be less if it abides by the capacity condition \eqref{eqn:capacity}, according to lemma~\ref{lemma:upper_bound_rechargeing_times}), which is a sign of overloading the
system. Reducing the robots to 4 gives
$\delta_{t_{cr}} = 41.1\si{\second}>\delta_t$ for
$\varepsilon = 0.27$. $E(t)$ and $E_{min}(t)$ are depicted in
Figs.~\ref{fig:volt2} and~\ref{fig:emin2}.
\subsection{Base scenario with wind and less $k_v$}
Here we consider the same previous scenario, but with robots having $k_v = 0.0045$. Here for 5 robots $\delta_{t_{cr}} = 48.3\si{\second}$ for $\varepsilon=0.14$, which alludes to the possibility of adding a robot. Indeed, for 6 robots $\delta_{t_{cr}} = 39.5\si{\second}>\delta_t$ for $\varepsilon = 0.14$. $E(t)$ and $E_{min}(t)$ are depicted in figures~\ref{fig:volt3} and~\ref{fig:emin3}. 
\section{Conclusions} 
\label{sec:conclusion}
In this paper we propose a CBF based framework for ensuring energy sufficiency of a multi-robot system while  sharing one charging station in a mutually exclusive manner, given that the robots are affected by air drag and known constant wind fields. 

As a future work we consider extending the current framework to allow for sharing multiple charging stations, as well as exploring the possibility of relaxing the assumption of having complete communication graph.

\bibliographystyle{IEEEtran}
\bibliography{references}

\begin{thebibliography}{10}
\providecommand{\url}[1]{#1}
\csname url@rmstyle\endcsname
\providecommand{\newblock}{\relax}
\providecommand{\bibinfo}[2]{#2}
\providecommand\BIBentrySTDinterwordspacing{\spaceskip=0pt\relax}
\providecommand\BIBentryALTinterwordstretchfactor{4}
\providecommand\BIBentryALTinterwordspacing{\spaceskip=\fontdimen2\font plus
\BIBentryALTinterwordstretchfactor\fontdimen3\font minus
  \fontdimen4\font\relax}
\providecommand\BIBforeignlanguage[2]{{%
\expandafter\ifx\csname l@#1\endcsname\relax
\typeout{** WARNING: IEEEtran.bst: No hyphenation pattern has been}%
\typeout{** loaded for the language `#1'. Using the pattern for}%
\typeout{** the default language instead.}%
\else
\language=\csname l@#1\endcsname
\fi
#2}}

\bibitem{portugal2011survey}
D.~Portugal and R.~Rocha, ``A survey on multi-robot patrolling algorithms,'' in
  \emph{Doctoral conference on computing, electrical and industrial
  systems}.\hskip 1em plus 0.5em minus 0.4em\relax Springer, 2011, pp.
  139--146.

\bibitem{cortes2004coverage}
J.~Cortes, S.~Martinez, T.~Karatas, and F.~Bullo, ``Coverage control for mobile
  sensing networks,'' \emph{IEEE Transactions on robotics and Automation},
  vol.~20, no.~2, pp. 243--255, 2004.

\bibitem{burgard2005coordinated}
W.~Burgard, M.~Moors, C.~Stachniss, and F.~E. Schneider, ``Coordinated
  multi-robot exploration,'' \emph{IEEE Transactions on robotics}, vol.~21,
  no.~3, pp. 376--386, 2005.

\bibitem{petersen2011termes}
K.~H. Petersen, R.~Nagpal, and J.~K. Werfel, ``Termes: An autonomous robotic
  system for three-dimensional collective construction,'' \emph{Robotics:
  science and systems VII}, 2011.

\bibitem{sun2003energy}
Z.~Sun and J.~Reif, ``On energy-minimizing paths on terrains for a mobile
  robot,'' in \emph{2003 IEEE International Conference on Robotics and
  Automation (Cat. No. 03CH37422)}, vol.~3.\hskip 1em plus 0.5em minus
  0.4em\relax IEEE, 2003, pp. 3782--3788.

\bibitem{slijepcevic2001power}
S.~Slijepcevic and M.~Potkonjak, ``Power efficient organization of wireless
  sensor networks,'' in \emph{ICC 2001. IEEE International Conference on
  Communications. Conference Record (Cat. No. 01CH37240)}, vol.~2.\hskip 1em
  plus 0.5em minus 0.4em\relax IEEE, 2001, pp. 472--476.

\bibitem{sonia}
A.~Kwok and S.~Martinez, ``Energy-balancing cooperative strategies for sensor
  deployment,'' in \emph{2007 46th IEEE Conference on Decision and
  Control}.\hskip 1em plus 0.5em minus 0.4em\relax IEEE, 2007, pp. 6136--6141.

\bibitem{ding2019decentralized}
Y.~Ding, W.~Luo, and K.~Sycara, ``Decentralized multiple mobile depots route
  planning for replenishing persistent surveillance robots,'' in \emph{2019
  International Symposium on Multi-Robot and Multi-Agent Systems (MRS)}.\hskip
  1em plus 0.5em minus 0.4em\relax IEEE, 2019, pp. 23--29.

\bibitem{notomista2018persistification}
G.~Notomista, S.~F. Ruf, and M.~Egerstedt, ``Persistification of robotic tasks
  using control barrier functions,'' \emph{IEEE Robotics and Automation
  Letters}, vol.~3, no.~2, pp. 758--763, 2018.

\bibitem{fouadEnergy2020}
H.~Fouad and G.~Beltrame, ``Energy autonomy for resource-constrained multi
  robot missions,'' in \emph{International Conference on Intelligent Robots and
  Systems (IROS)}, 2020.

\bibitem{ames2019control}
A.~D. Ames, S.~Coogan, M.~Egerstedt, G.~Notomista, K.~Sreenath, and P.~Tabuada,
  ``Control barrier functions: Theory and applications,'' in \emph{2019 18th
  European Control Conference (ECC)}.\hskip 1em plus 0.5em minus 0.4em\relax
  IEEE, 2019, pp. 3420--3431.

\bibitem{ames2016control}
A.~D. Ames, X.~Xu, J.~W. Grizzle, and P.~Tabuada, ``Control barrier function
  based quadratic programs for safety critical systems,'' \emph{IEEE
  Transactions on Automatic Control}, vol.~62, no.~8, pp. 3861--3876, 2016.

\bibitem{xu2015robustness}
X.~Xu, P.~Tabuada, J.~W. Grizzle, and A.~D. Ames, ``Robustness of control
  barrier functions for safety critical control,'' \emph{IFAC-PapersOnLine},
  vol.~48, no.~27, pp. 54--61, 2015.

\bibitem{xiao2019control}
W.~Xiao and C.~Belta, ``Control barrier functions for systems with high
  relative degree,'' in \emph{2019 IEEE 58th Conference on Decision and Control
  (CDC)}.\hskip 1em plus 0.5em minus 0.4em\relax IEEE, 2019, pp. 474--479.

\bibitem{li2018formally}
A.~Li, L.~Wang, P.~Pierpaoli, and M.~Egerstedt, ``Formally correct composition
  of coordinated behaviors using control barrier certificates,'' in \emph{2018
  IEEE/RSJ International Conference on Intelligent Robots and Systems
  (IROS)}.\hskip 1em plus 0.5em minus 0.4em\relax IEEE, 2018, pp. 3723--3729.

\bibitem{wang2016multi}
L.~Wang, A.~D. Ames, and M.~Egerstedt, ``Multi-objective compositions for
  collision-free connectivity maintenance in teams of mobile robots,'' in
  \emph{2016 IEEE 55th Conference on Decision and Control (CDC)}.\hskip 1em
  plus 0.5em minus 0.4em\relax IEEE, 2016, pp. 2659--2664.

\bibitem{egerstedt2018robot}
M.~Egerstedt, J.~N. Pauli, G.~Notomista, and S.~Hutchinson, ``Robot ecology:
  Constraint-based control design for long duration autonomy,'' \emph{Annual
  Reviews in Control}, vol.~46, pp. 1--7, 2018.

\end{thebibliography}
\clearpage


\begin{appendices}
\section{Proof of theorem \ref{thm:he_hocbf}}
\begin{proof}
	Since $u\in\mathbb{R}^2$ then there should be a value of control input that satisfies \eqref{eqn:he_ineq} provided that $L_gL_fh(x)\neq 0$. $L_gL_fh(x)$ can be written as 
	\begin{equation}
	L_gL_fh(x) = -kv\frac{(\mathbf{v} - \mathbf{v}_w)^T}{||\mathbf{v} - \mathbf{v}_w||}-\frac{k_c}{D}\frac{(\mathbf{x} - \mathbf{x}_c)^T}{D}
	\end{equation}
	both expressions are basically unit vectors multiplied by an expression or a factor. If we want the second expression to dominate the first one (so even if both vectors are opposite, the summation will not be equal to zero), we pick $k_c$ so that the least possible value of $\frac{k_c}{D}$ be greater than $k_v$ so $\frac{k_c}{R_0}>k_v\Rightarrow k_c > R_0 k_v$, meaning $L_gL_fh(x)\neq 0$.
\end{proof}

\section{Proof of lemma \ref{lemma:emin_arrival}}
\begin{proof}
	To show this, it is useful to point out to the fact that the minimum value of the quadratic cost function of the QP \eqref{eqn:QP} would be $u = u_{nom}$ in case $u_{nom}$ doesn't violate the constraints on the QP. Otherwise, the QP produces a value of $u$ that abides with the constraint in the equality sense (produces a control input that renders the constraint as an equality).
	
	Provided that the system starts in the safe set for $h_e > 0$, then at some time $T_b$ the nominal control action will cause inequality \eqref{eqn:he_ineq} to be violated, in which case the QP produces a safe control input $u$ that follows the constraint in the sense of equality. Therefore the produced control input causes $h_e$ to vary in the following way
	\begin{equation}
	\ddot{h}_e+(p_1+p_2)\dot{h}_e+p_1p_2h_e = 0
	\end{equation}
	for which the solution is 
	\begin{equation}
	h(t) = Ae^{\lambda_1t}+Be^{\lambda_2t}
	\end{equation}
	where $\lambda_1=\frac{1}{2}\left(-(p_1+p_2)+|p_1-p_2|\right)$ (the dominant mode),$\lambda_2=\frac{1}{2}\left(-(p_1+p_2)-|p_1-p_2|\right)$, and the constants $A$ and $B$ are determined from the initial conditions on $h_e$ and $\dot{h}_e$ at the time $T_b$.
	
	When the robot arrives on the boundary of the charging station at time $t_a$ we have
	\begin{equation}
	h_e(t_a) = E(t_a)-E_{min}= Ae^{\lambda_1t_a}+Be^{\lambda_2t_a}
	\end{equation}
	thus by properly choosing $p_1$ and $p_2$ we can gauge how closely the robot tracks $E_{min}$ on arrival to the charging station. We also point out to the fact that $h_e = 0$ only at the boundary of the charging region, because if $E=E_{min}$ (which is the boundary of our safe set so to speak) we want this to be at $h_e = 0$, which happens if $\log\frac{D}{\delta}=0 \Rightarrow D = \delta$.
	
	Since $h_e$ is a HOCBF, then any control input satisfying \eqref{eqn:he_ineq} renders the safe set forward invariant ($h_e\geq 0$) so in case if $h_e = 0$ and being on the boundary at the same time, then $E-E_{min}=0$.
\end{proof}

\section{A method for choosing $k_c$}
In this discussion we provide a heuristic to choose the value of $k_c$ in the definition of the energy sufficiency CBF. The third term in the definition of $h_e$ signifies the voltage change that a robot needs to go back to the charging station \cite{notomista2018persistification}. The basic idea of choosing $k_c$ starts by supposing that a robot can use a PD controller to go back to the charging station, starting on the boundary of the operating range (i.e. $||\mathbf{x}_0||=R_0$). For more conservatism, we suppose that there is a headwind with a magnitude of $||\mathbf{v}_w||$ opposing the robot's motion. Without loss of generality, we suppose that the robot is moving on a line so the robot's motion is 1-D, and that the charging station is in the origin. In this case the system's model will be
\begin{equation}
\begin{split}
\dot{x} &= v\\
\dot{v} &= -k_px-k_dv-C_d(v-||\mathbf{v}_w||)\\
\end{split}
\end{equation}
which is a second order ordinary differential equation, the solution of which is \footnote{The solution has been obtained using symbolic manipulation in Matlab.}
\begin{equation}
\begin{split}
x(t) &= \frac{R_0}{G}\left[\left(L_2-c\right)e^{-L_2t}-\left(L_1-c\right)e^{-L_1t}\right]\\&+\frac{C_d||\mathbf{v}_w||}{Gk_p}\left[L_1\left(e^{-L_2t}-1\right)-L_2\left(e^{-L_1t}-1\right)\right]\\
v(t) &= \frac{C_d||\mathbf{v}_w||-R_0k_p}{D}\left(e^{-L_1t}-e^{-L_2t}\right)
\end{split}
\end{equation}
where $c = k_d+C_d$, $G = \sqrt{c^2-4k_p}$,$L_1 = \frac{c-G}{2}$,$L_2 = \frac{c+G}{2}$. We then approximate the time needed to go from the initial position to a distance $\delta$ from the center (arriving at the boundary of the charging region) by taking only the dominant terms in consideration, so the position equation will be
\begin{equation}
x(t) = \left(\frac{R_0}{G}(c-L_1)-\frac{C_d||\mathbf{v}_w||}{Gk_p}L_2\right)e^{-L_1t}+\frac{C_d||\mathbf{v}_w||}{Gk_p}L_2
\end{equation}
thus the time at which $x(t)=\delta$ is
\begin{equation}
\Delta T = \frac{1}{L_1}\log\left(\frac{-C_d||\mathbf{v}_w||L_2+(c-L_1)k_pR_0}{ -C_d||\mathbf{v}_w||L_2+k_pG\delta }\right)
\end{equation}
then in order to consider the voltage change during this trip back to the charging station, we can integrate the voltage rate $\dot{E} = -k_e-k_v||v-v_w||$, however, to increase the conservatism in the estimate we choose to consider that the robot is moving on a constant speed equal to the maximum peak speed of $v(t)$, which can be obtained by differentiating $v(t)$ and getting the time at which the differential is equal to zero and use the velocity at this time, and we call it $v^*$ which is expressed as
\begin{equation}
v^* = \frac{-R_0k_p+C_d||\mathbf{v}_w||}{G}\left(\left(\frac{L_2}{L_1}\right)^{\frac{-L_2}{L_2-L_1}}-\left(\frac{L_2}{L_1}\right)^{\frac{-L_1}{L_2-L_1}}\right)
\end{equation}
and the voltage change needed becomes 
\small
\begin{equation}
\begin{split}
&\Delta E = \dot{E}\Delta T\\& = -\underbrace{\frac{k_e+k_v(|v^*| + ||\mathbf{v}_w||)}{L_1}}_{k_c}\log\left(\frac{-C_d||\mathbf{v}_w||+(c-L_1)k_pGR_0}{ -C_d||\mathbf{v}_w||+k_pG\delta }\right)
\end{split}
\end{equation}
\normalsize
however the current form of $\Delta E$ does not necessarily satisfy the condition that $E = E_{min}$ only on the boundary of the charging region, so we choose 
\begin{equation}
\Delta E = -k_c\log\left(\frac{D}{\delta}\right)
\end{equation}

\section{Proof of Theorem~\ref{thm:hc}}

\begin{proof}
	Since $\eta\in\Theta=\mathbb{R}$ there exists a control action $\eta$
	that satisfies~\eqref{eqn:coord_constraint}) (and keeps $\mathcal{C}_{ij}$
	invariant), then to show that $h_{c_{ij}}$ is a ZCBF, we need to ensure 	that $|T_{L_{i}}-T_{L_{j}}|\neq0$.
	
	The only chance that this difference can be equal to zero is when one of the
	robots enters to the charging region. To show this, consider having two
	robots $(i,j)$ applying~\eqref{eqn:coord_constraint}and without	loss of generality suppose that robot $j$ arrives at the charging station,
	so the difference in arrival times is
	\begin{equation}
	\Delta T_{L_{ij}} = \frac{E_i-E_{min_i}}{k_e+k_v\bar{V}_i}-\frac{E_j-E_{min_j}}{k_e+k_v\bar{V}_j} > 0
	\label{eqn:malcom_in_the_middle}
	\end{equation}
	it suffices to show that by the end of the charging process $\Delta T_{ij}<0$ which means that $\Delta T_{ij}=0$ at some point. To show this, we first point out that when the QP manipulates $\eta$ for coordination, then \eqref{eqn:coord_constraint} becomes an equality, and due to the choice of $\gamma$ in \eqref{eqn:gamma_choice}, the right hand side will be equal to zero. Thus
	\begin{equation}
	\eta_i\approx-(k_e+k_v\bar{V}_i)\left(-\frac{k_e+k_vV_j}{k_e+k_v\bar{V}_j}+\frac{k_e+k_vV_i}{k_e+k_v\bar{V}_i}\right)
	\label{eqn:eta_approx}
	\end{equation}
	notice that we neglected the expression $\frac{k_v}{w}\frac{(E-E_{min})(V(t)-V(t-w))}{k_e+k_v\bar{V}}$ because it can be significantly less than $\frac{k_e+k_vV_i}{k_e+k_v\bar{V}_i}$ for most cases of $w$. An extreme case for \eqref{eqn:eta_approx} can be anticipated if we neglect $\frac{k_e+k_vV_j}{k_e+k_v\bar{V}_j}$ alltogether and take $V_i=\max(V_n,\tilde{V})$, so an extreme case for $\eta_i$ can be
	\begin{equation}
	\eta_i \approx -(k_e+k_v\max\{V_n,\tilde{V}\})
	\end{equation}
	however if $|\eta_i|$ is less than $k_{ch}$ (by assumption) it means that  $E_j$ increases faster than $E_{min}$ changes for both robots $(i,j)$ (noticing that for $\eta_j$ the bracket in \eqref{eqn:eta_approx} will be of reversed sign).  Without loss of generality, for cases where $k_{ch} \gg(k_e+k_v\max\{V_n,\tilde{V}\})$ we can consider the change in $E_{min}$ values is sufficiently slow that they can be considered constant, so from \eqref{eqn:malcom_in_the_middle} the value of $T_{L_j}$ is increasing in a faster rate than $T_{L_i}$ is decreasing, and at some point $T_{L_j} = \frac{E_{max}-E_{min_j}}{k_e+k_v\bar{V}_j} > \frac{E_i-E{min_i}}{k_e+k_v\bar{V}_i}=T_{L_i}$, which means that $\Delta T_{L_{ij}}=0$ at some point during the recharge.

	The proof of the second part is the same as that of proposition III.1
	in~\cite{li2018formally} and is omitted for brevity.
\end{proof}
\begin{remark}
	To demonstrate the fact that $k_{ch}$ is sufficiently big, a minimum threshold on $k_{ch}$ can be obtained by equating $\delta_{t_{cr}}$ with $\frac{E_{max}-E_{lb}}{2}$ in \eqref{eqn:capacity}. In other words, by doing so we can get a minimum acceptable value of $k_{ch}$ so that the capacity constraint \eqref{eqn:capacity} is technically satisfied. Doing so we get
	\begin{equation}
	k_{ch} = (k_e+k_v\tilde{V})\left[\tfrac{(2(n-1)\Delta E+\varepsilon) \pm \sqrt{(2(n-1)\Delta E+\varepsilon)^2+4(n-1)(\Delta E-2\varepsilon)\Delta E}}{2(\Delta E-2\varepsilon)}\right]
	\end{equation}
	where $\Delta E = E_{max}-E_{lb}$. If we set $\varepsilon=0$ for simplicity, and for $n=2$ we get $k_{ch}=(1+\sqrt{2})(k_e+k_v\tilde{V})$. In practice, the value of $\delta_{t_{cr}}$ is usually significantly larger than $\frac{E_{max}-E_{lb}}{2k_{ch}}$. Moreover the value of $n$ is bigger than two, which means that $k_{ch}$ is in practice significantly larger than $(k_e+k_v\tilde{V})$. 
\end{remark}
\section{Proof of lemma \ref{lemma:upper_limit_recharge}}
\begin{proof}
	We start by showing that 
	\begin{equation*}
	\bar{E}_M \leq \frac{E_{max}+E_{lb}}{2}
	\end{equation*}
	To do this, we calculate the difference 
	\begin{equation}
	\begin{split}
	&\tfrac{E_{max}+E_{lb}}{2}-\bar{E}_M = \tfrac{E_{max}+E_{lb}}{2}-\tfrac{(1+\tfrac{k_e+k_v\tilde{V}}{k_{ch}})E_{max}+E_{lb}-\delta_t(k_e+k_v\tilde{V})}{\kappa}
	\end{split}
	\end{equation}
	where $\kappa = 2+\tfrac{k_e+k_v\tilde{V}}{k_{ch}}$. This gives
	\begin{equation}
	\tfrac{E_{max}+E_{lb}}{2}-\bar{E}_M =\tfrac{(k_e+k_v\tilde{V})}{2\kappa}\left(2\delta_t -\tfrac{E_{max}-E_{min}}{k_{ch}}\right)
	\end{equation}
	but due to the choice \eqref{eqn:lower_bnd_dt} then $\tfrac{E_{max}+E_{lb}}{2}-\bar{E}_M \geq 0$
	This sets an upper bound on the value of $E_{min}$ of the most needy agent (with which it arrives to the charging station). Now the  number of arrivals of a robot in a cycle is 
	\begin{equation}
	\zeta_i= 1+\left\lfloor \frac{\tfrac{(E_{max}-E_{lb})\left(1+\tfrac{k_e+k_v\tilde{V}}{k_{ch}}\right)}{k_e+k_v\tilde{V}}}{\tfrac{(E_{max}-\bar{E}_{M})\left(1+\tfrac{k_e+k_v\tilde{V}}{k_{ch}}\right)}{k_e+k_v\tilde{V}}}\right\rfloor
	\end{equation}
	where $\lfloor. \rfloor$ is the floor operator. Here the numerator represents the time the least needy robot (that defines the cycle) takes to discharge and recharge once, while the denominator expresses the same thing for the most needy robot. The ratio represents how many whole sections to which a cycle can be divided, or in other words, how many small cycles can we fit in the large one (i.e. how many visits the most needy robot can do in a cycle). Since we are considering the case where $\bar{V}=\tilde{V}$ for all robots to be more conservative, then $\zeta_i$ can be reduced to 
	\begin{equation}
	\zeta_i = 1 + \left\lfloor\frac{E_{max}-E_{lb}}{E_{max}-\bar{E}_{M}}\right\rfloor
	\end{equation}
	substituting the upper bound of $\bar{E}_M$ in the last equation
	\begin{equation}
	\begin{split}
	\zeta_i &= 1+\left\lfloor\frac{E_{max}-E_{lb}}{E_{max}-\tfrac{E_{max}+E_{lb}}{2}}\right\rfloor=1+\left\lfloor\frac{E_{max}-E_{lb}}{E_{max}-E_{lb}}\right\rfloor\\
	&= 2
	\end{split}
	\end{equation}
	since this has been considered for the most needy robot, this means that all other robots, which have less $E_{min}$ values, visit the charging station at most two times per cycle. Notice that in this proof we used the more critical value of $\bar{E}_M$ at which the robot arrives to the charging station.
\end{proof}

\section{Proof of lemma \ref{lemma:feasibility}}
\begin{proof}
	Since from lemma \ref{lemma:upper_limit_recharge} we know that for any robot the maximum number of visits to the charging station is at most two, then the maximum number of total visits to the charging station \emph{within} one cycle is $2(n-1)$. Moreover, the total number of spaces between these visits (taking the start and end of the cycle into account) is $M = 2(n-1)+1 = 2n-1$. We then calculate the amount of available time between visits $\delta_{av}$ by dividing the cycle length (while still assuming that all robots operate such that $\bar{V}=\tilde{V}$) and compare this quantity to $\delta_{t_{cr}}$
	\begin{equation}
	\delta_{av} = \frac{(E_{max}-E_{lb})(1+\tfrac{k_e+k_v\tilde{V}}{k_{ch}})}{(2n-1)(k_e+k_v\tilde{V})}
	\end{equation}
	To check that $\delta_{av} > \delta_{t_{cr}}$ we calculate the difference
	\begin{equation}
	\begin{split}
	\delta_{av}-\delta_{t_{cr}}&=\tfrac{(E_{max}-E_{lb})(1+\tfrac{k_e+k_v\tilde{V}}{k_{ch}})}{(k_e+k_v\tilde{V})}\left(\tfrac{1}{(2n-1)}-\tfrac{1}{1+\kappa(n-1)}\right)\\&+\tfrac{\kappa\varepsilon}{k_e+k_v\tilde{V}}
	\end{split}
	\end{equation}
	but since $1+1+\kappa(n-1) = 2n-1 + \tfrac{k_e+k_v\tilde{V}}{k_{ch}}(n-1)>2n-1$, and that $\kappa\varepsilon>0$, then $\delta_{av} - \delta_{t_{cr}}>0$, meaning that the available time is bigger than $\delta_{t_{cr}}$, which means a $\delta_t$ satisfying \eqref{eqn:capacity} can be accommodated(since $\delta_{t_{cr}}$ can be accommodated).
\end{proof}

\section{Proof of Theorem ~\ref{thm:main_result}}
\begin{proof}
	~\cite[Theorem 3]{fouadEnergy2020}
	From Algorithm~(\ref{algo}), each robot is either applying the coordination
	CBF $h_{c_{ij}}$ or the lower bound CBF $h_L$. For the robots which don't
	apply $h_L$, the value of the control input $\eta_i$ that respects \eqref{eqn:coord_constraint} leads
	$E_{min_i}$ into safe set $\mathcal{C}_{ij}$ with respect to its neighbour with the closest landing time (by virtue of theorem ~\ref{thm:hc}). Each robot applies this to its neighbour with
	the closest landing time
	$\{(i,j) | j\in\mathcal{N}_i\text{ and }
	h_{c_{ij}}=\min_{k\in\mathcal{N}_i}h_{c_{ik}}\}$, eventually leading to
	$E_{min_i}\in\mathcal{C}=\bigcap\limits_{\forall i \neq
		j}\mathcal{C}_{ij}\quad ,\forall i$.
	Moreover, since we have established the feasibility of the scheduling
	problem in Lemma~(\ref{lemma:feasibility}), then we know that the sets
	$\mathcal{C}_{ij}$ are nonempty and that a solution exists.
	
	If a robot $i$ is applying the lower bound $h_L$, then it can't push its
	arrival time any further. In this case The nearest robot $j$ that applies
	the coordination CBF will have a control action $\eta_j$ that will lead
	$E_{min_j}$ to $\mathcal{C}_{ij}$ (noticing that $\mathcal{C}_{ij}$ is non
	empty), and then all other robots applying coordination CBF will coordinate
	in a pairwise fashion based on the neighbour of closest landing time as
	discussed in the previous point. If we add to the previous points the ability of each robot to arrive at the charging station at almost $E_{min}$ (by virtue of lemma \ref{lemma:emin_arrival}), then mutual exclusive use of the charging station is satisfied.	
\end{proof}
\section{Estimation of $\varepsilon$ parameter}
To motivate the need for  $\varepsilon$ , let's consider a pair of consecutive robots in the charging schedule which are manipulating $\eta_i$ 
so that their values of $E_{min}$ stay inside $\mathcal{C}_{ij}$ or on its boundary. We are interested in the critical case when \eqref{eqn:coord_constraint} is violated (when both $E_{min}$ values start outside $\mathcal{C}_{ij}$ or approach to the boundary from the inside), in which case the QP produces values of $\eta$ that renders \eqref{eqn:coord_constraint} an equality, hence $\dot{h}_{c_{ij}}=-\alpha(h_{c_{ij}})$, which reaches a steady state in finite time~\cite{li2018formally}, i.e. $\dot{h}_{c_{ij}}=0$. Thus $\eta$ in \eqref{eqn:coord_constraint} changes such that $\dot{h}_{c_{ij}}=0$ after reaching the steady state. Supposing that both robots operate on the maximum nominal speed of the mission $V_n$(relative w.r.t. wind)  such that they have an equal average relative speed w.r.t. wind $\bar{V}_i=\bar{V}_j$, then as $\dot{h}_{c_{ij}}=0$, from LHS of \eqref{eqn:coord_constraint} we have $\eta_i = k_v(V_j-V_i)$. Suppose that robot $i$ goes back to the charging station and that its speed decreases exponentially from $V_n$ to $V_f\ll V_n$ with a rate $a$, then
\begin{equation}
\dot{E}_{min}=\eta = k_vV_n(1-e^{-at})
\label{eqn:inc_emin}
\end{equation}
which means that as the exponential term decreases, $\eta$ increases and thus $E_{min}$ increases. In order to be able to estimate this increase, we need to integrate \eqref{eqn:inc_emin} from the time the robot starts moving towards the charging station till it arrives.

\begin{figure}[!htb]
	\centering
	\includegraphics[width=\columnwidth]{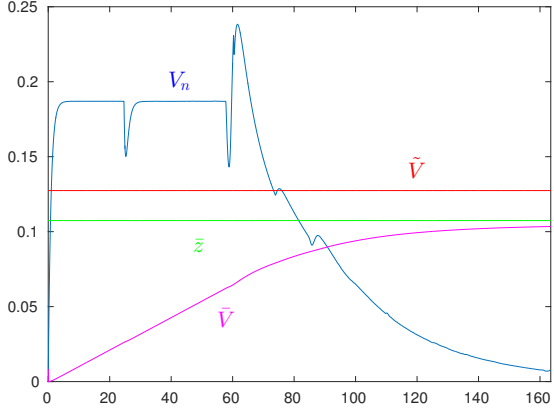}
	\caption{Demonstration of the maximum average relative velocity w.r.t. wind $\tilde{V}$, the average velocity $\bar{V}$ and the nominal mission velocity $V_n$ for a robot revolving around the charging station.}
	\label{fig:vel_app}
\end{figure}

Considering the most needy robot as this robot $i$\footnote{Since we considered $\bar{E}_M=E_M+\varepsilon$ for the most needy agent and defined $\delta_{t_{cr}}$ based on that}, then we can say that the arrival time $T_{end}$ in the most critical case is
\begin{equation}
T_{end}=\frac{E_{max}-\bar{E}_M}{k_e+k_v\tilde{V}}=\frac{E_{max}-\tfrac{E_{max}+E_{min}}{2}}{k_e+k_v\tilde{V}}
\label{eqn:t_end}
\end{equation}
{\color{black}notice here that for agent $i$ in the above equation, the average speed expression may include the mission segment (at which the robot operates at a speed equal to $V_n$), and the approach where the speed decreases, so for conservativeness we suppose that $\bar{V}_i=\tilde{V}$, which is the same thing we did on deriving $\delta_{t_{cr}}$}. An example demonstration for the aforementioned velocities is in Figure~\ref{fig:vel_app}.

To estimate the time at which the neediest robot starts approaching the charging station $T_{start}$, we can approximate it as being the time at which $h_e=0$ for this robot, while supposing it is operating at the boundary of the operating range $\mathcal{R}$\footnote{This approximation is based on the idea that the safe control input described by the constraints of the QP start taking over when the states of the system are close to the boundary of the safe set ($h_e=\epsilon$, where $\epsilon \ll$).}. This means
\begin{equation}
E(T_{start})-E_{min}(T_{start})-k_c\log\frac{R_0}{\delta}=0
\end{equation}
we can take $E(t) = E_{max}-(k_e+k_v\tilde{V})(t-t_0)$ where $t_0=0$ (considering the first cycle) and we can take $E_{min}(T_{start}) = E_M$. Substituting in the last equation we get
\begin{equation}
T_{start}=\frac{E_{max}-E_M-k_c\log\frac{R_0}{\delta}}{k_e+k_v\tilde{V}}
\end{equation}
substituting \eqref{eqn:em_ub} in the last equation we get
\begin{equation}
\begin{split}
T_{start}&=\frac{\varepsilon}{k_e+k_v\tilde{V}}\left(1-\frac{1}{1+\kappa(n-1)}\right)\\&+\frac{n(E_{max}-E_{lb})-k_c(1+\kappa(n-1))\log\tfrac{R_0}{\delta}}{(1+\kappa(n-1))(k_e+k_v\tilde{V})}
\end{split}
\label{eqn:t_start}
\end{equation}
Supposing that the robot decreases its velocity from $V_n$ to $v_f$ in an amount of time equal to $T_{end}-T_{start}$, then
\begin{equation}
\begin{split}
v_f= V_ne^{-a(T_{end}-T_{start})}
\end{split}
\end{equation}
and 
\begin{equation}
a = -\frac{1}{T_{end}-T_{start}}\log\frac{v_f}{V_n}
\end{equation}
Now in order to calculate the increase in $E_{min}$ 
\begin{equation}
\begin{split}
\varepsilon &= k_vV_n\int_{0}^{T_{end}-T_{start}}(1-e^{-at})dt\\
&=k_vV_n\left[t+\frac{1}{a}e^{-at}\right]_{0}^{T_{end}-T_{start}}\\
&=k_vV_n\left[T_{end}-T_{start}+\frac{1}{a}\left(\frac{v_f}{V_n}-1\right)\right]\\
&=k_vV_n\underbrace{\left[1+\frac{1}{\log\frac{V_n}{v_f}}\left(\frac{v_f}{V_n}-1\right)\right]}_{\Gamma}(T_{end}-T_{start})
\end{split}
\end{equation}
the smaller the choice of $v_f$, the closer $\Gamma$ approaches one, and the more conservative the estimate of $\varepsilon$ will be. Substituting \eqref{eqn:t_end} and \eqref{eqn:t_start} in the above equation we get
\begin{equation}
\varepsilon = \frac{\Gamma k_vV_n\left(T_{end}-\tfrac{n(E_{max}-E_{lb})-k_c(1+\kappa(n-1)\log\tfrac{R_0}{\delta}}{(k_e+k_v\tilde{V})(1+\kappa(n-1)}\right)}{1+\tfrac{\Gamma k_vV_n}{k_e+k_v\tilde{V}}\left(1-\tfrac{1}{1+\kappa(n-1)}\right)}
\end{equation}

\end{appendices}

\addtolength{\textheight}{-12cm}   



%

\end{document}